\documentclass[10pt,twocolumn,letterpaper]{article}

\usepackage{cvpr}

\usepackage{booktabs} 
\usepackage{amssymb}
\usepackage{pifont}
\usepackage{float}
\usepackage[dvipsnames]{xcolor}

\usepackage{gensymb}

\definecolor{revision_maroon}{rgb}{0.5, 0, 0}

\newcommand{\M}[0]{\mathbf{M}}
\newcommand{\G}[0]{\mathbf{G}}

\usepackage{xcolor}
\definecolor{myred}{rgb}{0.8,0,0}
\definecolor{mygreen}{rgb}{0,0.8,0}
\usepackage{pifont}
\newcommand{\cmark}{{\color{mygreen}\ding{51}}}
\newcommand{\xmark}{{\color{myred}\ding{55}}}
\DeclareMathOperator*{\argmin}{argmin}

\usepackage[ruled]{algorithm2e} 

\SetAlFnt{\small}
\SetAlCapFnt{\small}
\SetAlCapNameFnt{\small}
\SetAlCapHSkip{0pt}

\definecolor{cvprblue}{rgb}{0.21,0.49,0.74}
\usepackage[pagebackref,breaklinks,colorlinks,citecolor=cvprblue]{hyperref}

\title{\scalebox{0.95}{\hbox{ROAM: \textbf{R}obust and \textbf{O}bject-\textbf{A}ware \textbf{M}otion Generation Using Neural Pose Descriptors}}}

\author{
\;Wanyue Zhang\textsuperscript{1,2}\quad\quad
Rishabh Dabral\textsuperscript{1,2}\quad\;
Thomas Leimkühler\textsuperscript{1}\\
Vladislav Golyanik\textsuperscript{\textdagger,1}~~~
Marc Habermann\textsuperscript{\textdagger,1,2}~~~
Christian Theobalt\textsuperscript{1,2}
\smallskip\\
\textsuperscript{1}Max Planck Institute for Informatics, Saarland Informatics Campus~~~~~~\\
\textsuperscript{2}Saarbrücken Research Center for Visual Computing, Interaction and AI
}

\begin{document}

\maketitle
\vspace*{-0.6cm}%
\footnotetext[2]{Equal advising/contribution and corresponding authors}

\begin{abstract}
Existing automatic approaches for 3D virtual character motion synthesis supporting scene interactions do not generalise well to new objects outside training distributions, even when trained on extensive motion capture datasets with diverse objects and annotated interactions. 
This paper addresses this limitation and shows that robustness and generalisation to novel scene objects
in 3D object-aware character synthesis can be achieved by training a motion model with as few as one reference object. 
We leverage an implicit feature representation trained on object-only datasets, which encodes an $\mathsf{SE(3)}$-equivariant descriptor field around the object. 
Given an unseen object and a reference pose-object pair, we optimise for the object-aware pose that is closest in the feature space to the reference pose.
Finally, we use \textit{l}-NSM, \textit{i.e.,} our motion generation model that is trained to seamlessly transition from locomotion to object interaction with the proposed bidirectional pose blending scheme. 
Through comprehensive numerical comparisons to state-of-the-art methods and in a user study, we demonstrate substantial improvements in 3D virtual character motion and interaction quality and robustness to scenarios with unseen objects. Our project page is available at \url{https://vcai.mpi-inf.mpg.de/projects/ROAM/}.
\end{abstract}
%
%
\begin{figure}
    \centering
    \captionsetup{type=figure}
    \includegraphics[width=\linewidth]{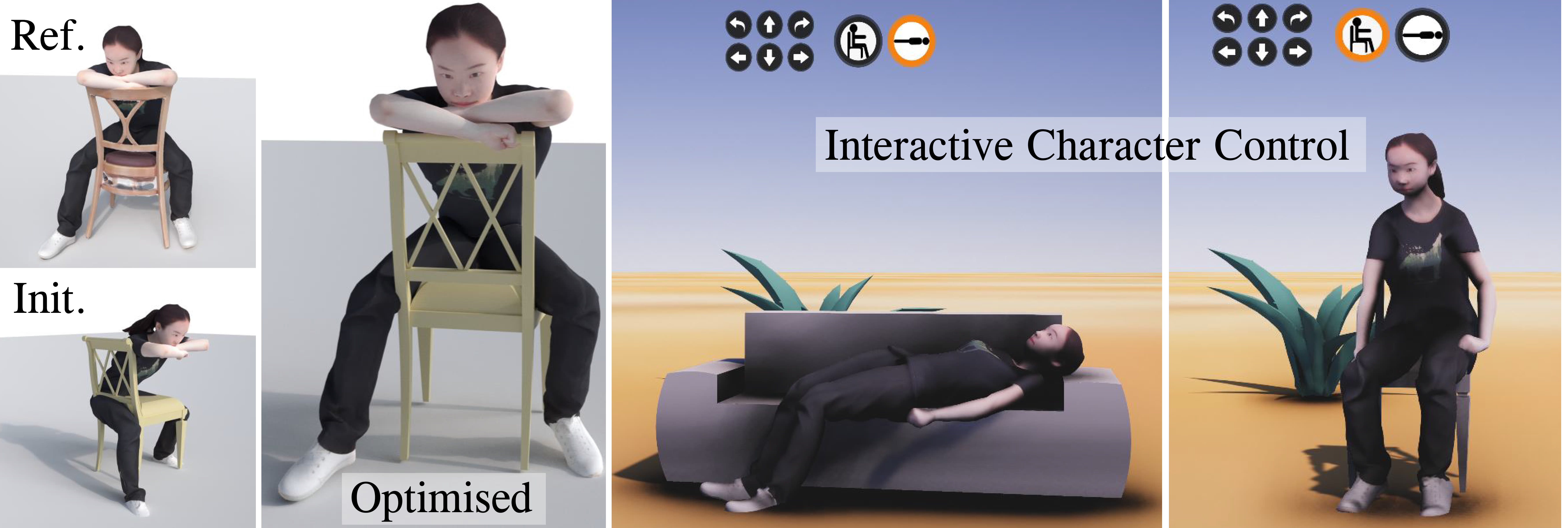}
    \captionof{figure}{
	ROAM generates sitting and lying poses on a large variety of chairs and couches. It is $\mathsf{SE(3)}$-equivariant and robust to out-of-distribution objects of the same category. 
    ROAM also allows interactive character control without requiring the model to be trained on mocap data with subjects interacting with diverse chairs and couches. 
    }
    \label{fig:teaser}
    \vspace{-0.6cm}
\end{figure}
%
%
%
\vspace{-0.3cm}
\section{Introduction} \label{sec:intro}
%
%
%
%

Data-driven virtual 3D character animation has recently witnessed remarkable progress, and several neural methods modelling human motion have been proposed~\cite{holden2017phase, starke2019neural, deepphase}. 
The realism of virtual characters is a core contributing factor to the quality of computer animations and user experience in immersive applications like games, movies, and VR/AR. 
In such settings, a virtual character often interacts with different assets in the environment (\textit{e.g.,} pieces of furniture) and elaborate interactions usually require manual work of professional users. 
However, automatically synthesising natural animations with scene interaction constraints (as shown in Fig.~\ref{fig:teaser}) remains unsolved and challenging, especially when the virtual assets in the scene significantly differ from those captured during the data acquisition phase. 
%
%
%
%
\par
Approaches available in the literature enable human motion and performance capture~\cite{luvizon2023, Shimada2022, liu20214d, weng2021holistic, chen2019holistic++, HPS, bhatnagar2022behave, Dabral_2021_ICCV}, 
scene-agnostic animation~\cite{10.1145/3528223.3530178, 8374600, dholden, yuan2020dlow, Aksan_2019_ICCV, wanghongsong, xujingwei, 9157765, pavllo:quaternet:2018, dabral2022mofusion, martinez17}, 
fine-grained grasping and manipulation synthesis~\cite{taheri2022goal, prokudin2019efficient, zhou2022toch, taheri2020grab, ghosh2022imos, tendulkar2023flex, zhang2024artigrasp, zheng2023cams}), interaction with dynamic objects~\cite{braun2023physically, xu2023interdiff, lee2023locomotion, li2023object, li2023controllable}, and inverse problems such as inferring object states from human interactions~\cite{petrov2023popup, 10.1145/3550469.3555426, jiang2023full}.
In our context, several existing works on scene-aware character animation focus on specific interaction types such as \textit{sitting} on a chair \cite{starke2019neural, deepphase, zhang2022couch, hassan2021stochastic}. 
Despite producing fairly natural animation for \textit{known} objects, such methods struggle to generalise to unseen objects of the same category.
Creating a generalisable motion model, which is capable of synthesising \textit{object-aware} and \textit{realistic} motions remains very challenging and unsolved. 
This is a major shortcoming due to the exponential number of possible asset-motion combinations.  
Na\"{\i}vely training a generalisable neural network to synthesise sitting and lying motion would require collecting and annotating a large dataset of object-interaction motions with a large variety of chairs and couches. 
This is expensive and laborious. 
We believe all these challenges necessitate a fundamentally new approach to character-environment interaction modelling. 
%
%
%
%
\par 
Hence, we propose ROAM, \textit{i.e.,} an alternative solution that works around the requirement of large and diverse datasets while also providing a complete framework for virtual character animation with object interaction; see Fig.~\ref{fig:overview} for an overview. 
ROAM is a robust and object-aware virtual character animation framework to synthesise human-object interactions; it generalises to unseen objects of the same category \emph{without} relying on a large dataset of human-object animations. 
To this end, we follow a \textit{divide-and-conquer} strategy to dataset collection, which utilises existing large-scale object-only dataset for goal pose generation and thus reduces the data demand for paired human-object interaction sequences. 
This affords us generalisability across instances of an object category while requiring character animation data with as few as \textit{one} exemplar for an object category. 
We divide the problem into two subtasks: \textit{goal pose estimation} and \textit{goal-driven motion synthesis}. 
\par 
Goal pose estimation is framed as a descriptor matching problem, which only requires a library of object shapes (such as ShapeNet \cite{chang2015shapenet} or ObjaVerse \cite{objaverse}) and a desired reference pose. 
By designing $\mathsf{SE(3)}$-invariant neural descriptors~\cite{simeonov2022neural}, we encode the geometric relations for human-object interaction. 
This allows us to optimise the reference pose for an unseen object by minimising the descriptor distance between the reference human-object pair and the pose for the target object.
\par
Once the goal pose for a previously unseen target object is optimised, our proposed \textit{l}-NSM auto-regressively animates the character's motion from a standing pose to the sitting and lying goal pose. 
Remarkably, \textit{l}-NSM is trained using motion capture data with only a single exemplar object per category. 
This separation of tasks makes ROAM robust to variations in unseen object styles and scales while also avoiding expensive data capture overheads. 
%
%
%
%
\par
In summary, our technical contributions are as follows: 
\begin{itemize}
    \item ROAM, a new approach for synthesising virtual character-object animations in 3D, which generalises to unseen objects of the same type while avoiding the capture of large motion-capture datasets;
    \item A divide-and-conquer strategy simplifying the learning problem while also increasing the robustness of the method with respect to unseen object instances;
    \item A novel, $\mathsf{SE(3)}$-equivariant neural descriptor objective to generate the goal pose and its integration into \textit{l}-NSM, producing realistic motions towards the goal pose (Sec.~\ref{sec:gps}). 
\end{itemize}
%
%
%
%
\par
To evaluate our design choices, we record a dataset of sitting motion on a reference chair and a reference sofa as well as a lying down motion on a reference sofa, which we will make publicly available for future research.
Through a comprehensive set of experiments and perceptual study, we show the effectiveness of our method in terms of motion plausibility and robustness to unseen chair and sofa types.
%
%
%
\section{Related Work} \label{sec:related}
We now discuss two closely-related method classes, \textit{i.e.,} methods that generate \textit{static} poses conditioned on the scene (Sec.~\ref{ssec:RW_static}) and methods supporting \textit{dynamic} and scene-aware animations (Sec.~\ref{ssec:RW_dynamic}). 
Tab.~\ref{tab:concecptual_comparison} provides a conceptual comparison of our method to previous works. 
%
%
%
\subsection{Scene-aware Human Pose Generation}\label{ssec:RW_static} 
%
%
%
\begin{table} 
    \small 
	\begin{center} 
        \vspace{-0.2cm}
		\begin{tabular}{lccc}
            \toprule
             & w/o Man. Label       & Robust      &  Single Object         \\

            \midrule
            NSM \cite{starke2019neural}  		             & \xmark	        & \xmark	    & \xmark	        \\
	        SAMP \cite{hassan2021stochastic}  	                & \cmark	        & \xmark	    & \xmark	        \\	
            COUCH \cite{zhang2022couch}  		             	 & \xmark           & \xmark	    & \xmark	        \\
			\textbf{Ours}  			                          & \cmark           & \cmark        & \cmark            \\

            \bottomrule
		\end{tabular}
  		\caption
    	{
    	Closely related works in scene-aware 3D human pose and motion synthesis. 
    	\textit{Manual labelling} means that objects have to be manually annotated (\textit{e.g.,} contact points) at inference time.
    	\textit{Robust} approaches can account for various unseen instances of the same object category.
    	Finally, \textit{single object} refers to the requirement of motion capture data with \textit{one} object instance, \textit{i.e.,} a single chair. 
    	}
        \label{tab:concecptual_comparison}
        \vspace{-0.6cm}
	\end{center}
\end{table}
%
%

%
This setting requires generating a plausible 3D pose of a character, given the scene geometry. 
In this regard, \citet{zhang2020generating} use a conditional Variational Auto-Encoder (cVAE) to estimate a pose conditioned on the latent space of the scene and further optimise it to achieve physical plausibility. 
~\citet{li2019putting} build a large-scale dataset with scene affordance and use it to train a 3D generative model, which plausibly places humans in a scene. 
POSA~\cite{hassan2021populating} encodes contact probabilities between body mesh vertices and scene semantics. 
COIN~\cite{zhao2022compositional} synthesises compositional human-object interaction given object semantics and intended action while requiring extensive labelling of objects and interactions on the PROX dataset~\cite{hassan2019resolving}.
In contrast to existing methods, we introduce a different paradigm for object-aware human pose generation.
We also differ in that we are interested in synthesising dynamic motion sequences which can be controlled by a user.
%
%
\subsection{Scene-aware Human Motion Generation}\label{ssec:RW_dynamic} 
Dynamic, scene-aware motion generation requires a motion synthesis pipeline such that the characters perform a desired interaction with the scene.
Towards this goal, Neural State Machine (NSM)~\cite{starke2019neural} synthesises periodic and non-periodic human motion with scene interactions. 
{It encodes the environment and the objects using a coarse volumetric representation which can often lack robustness for unseen objects}.
However, it is not $\mathsf{SE(3)}$-equivariant and therefore requires the user to {\textit{manually label}} the goal position and orientation for the character to interact with the object. 
SAMP~\cite{hassan2021stochastic} improves upon NSM by predicting the goal location. 
However, it also uses the voxel-based scene representation, which limits the method's robustness and generalisability to novel instances of the same object category.
COUCH~\cite{zhang2022couch} studies human--chair interaction by first sampling hand contact points on the chair using a VAE and then synthesising motion for the hands to reach the regressed target locations. 
This approach only allows partial variability and controllability as it provides no constraints to joints that are not in contact with the object. 
Moreover, it relies on a large dataset with labelled contact points and lacks generalisability to unseen objects. 
\citet{wang2021synthesizing} tackles long-duration motion synthesis by first generating intermediate static poses and then synthesising short in-between motions. 
Scene constraints such as collision avoidance and affordance are incorporated during post-processing, prohibiting real-time performance. 
\par 
We argue that previous methods on 3D human pose and motion generation either tend to be bound to object instances close to the ones seen during training~\cite{starke2019neural} or rely on large and expensive data collections~\cite{zhang2022couch,hassan2021stochastic}. 
In contrast, our method does not rely on 
extensive motion capture sequences with diverse objects while still achieving robust motion synthesis. 
%
%
%
\section{Method} \label{sec:method}
\begin{figure}[!t]
	\includegraphics[width=\linewidth]{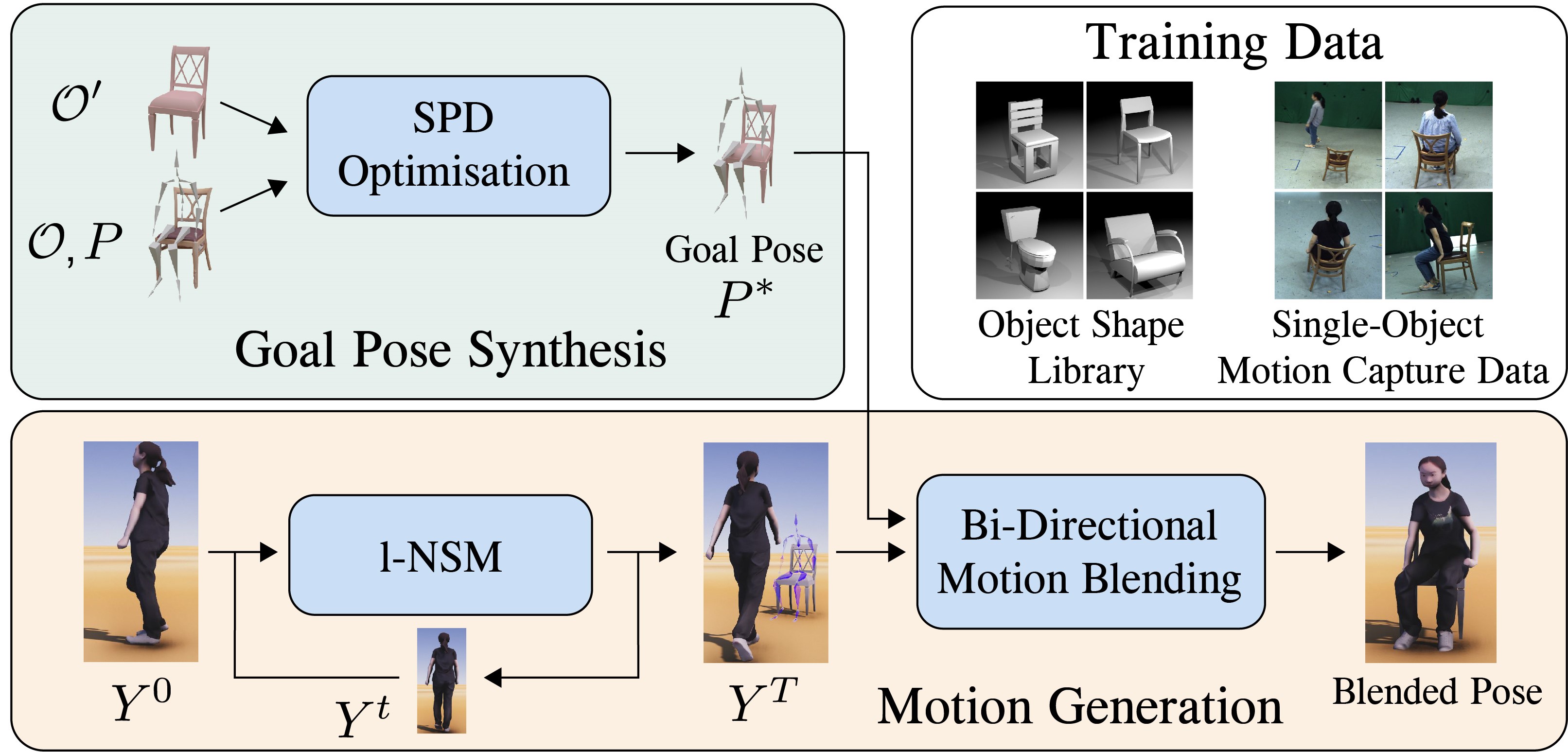}
	\caption
	{
	    \textbf{Method overview.} 
	    Given a starting standing pose, a sitting pose on a reference chair as well as a novel object (not seen during training), our goal is to synthesise a character animation, which approaches the novel object (either chair or sofa) and then sits or lies on it while the motion should be aware of the shape of the novel object.
	    We achieve this by our proposed three-stage procedure.
	    First, we propose a skeletal pose field, which adapts the sitting pose on the reference chair to the novel chair in a shape-aware manner while not requiring any manual labelling.
	    We call this pose the goal sitting pose.
	    Second, \textit{l}-NSM generates an approaching and interaction motion towards the novel object.
	} 
	\label{fig:overview}
 \vspace{-0.6cm}
\end{figure}
Our goal is to synthesise 3D motion sequences of a character approaching and interacting with chairs and couches of various geometries, scales, and orientations.
To learn such motions, we restrict ourselves to using motion-capture data with only one reference exemplar object per category; chairs and couches are considered separate categories.
The overall schema of our method is shown in Fig.~\ref{fig:overview}. 
We propose a \textit{divide-and-conquer} strategy consisting of two main stages:
First, our Goal Pose Synthesis (GPS) module takes a reference pose and an unseen target object and then generates an adapted goal pose, which fits the shape of the target object.
We do so using our novel Skeletal Descriptor Fields (Sec.~\ref{sec:gps}) which help the optimiser discover correspondences between the reference and the target object.
With the goal pose synthesised, we need to seamlessly integrate it into our motion synthesis pipeline for low-level control.
This is facilitated by our second component, lightweight Neural State Machine(\textit{l}-NSM), that synthesises a natural walking motion approaching the object and interacting with it from the starting pose towards the final goal pose (Sec.~\ref{sec:motion_net}).
Finally, we propose an improved bidirectional blending scheme that seamlessly integrates the \textit{l}-NSM motions with the goal pose.
\par
The key motivation behind this two-stage design is that each task in separation can be trained with less data. 
More precisely, the goal pose synthesis solely requires an object dataset and \textit{l}-NSM only requires motion-capture data paired with as few as one object per category. 
%
%
%
%
\subsection{Object-aware and Robust Goal Pose Synthesis} \label{sec:gps}
To estimate the goal pose that generalises to arbitrary shapes belonging to the same category, we first create a dataset of a character interacting with a reference object. 
Specifically, the dataset consists of several sequences of human  motion, $\mathcal{P} = \{P_1, P_2, \dots\}$, where $P_i$ is the character's pose at frame $i$.
We parameterise the pose using axis-angles, \textit{i.e.,} $P = \{\mathbf{t}_\mathrm{root}, \mathbf{p}_j\}$ for $j = \{1,\hdots,J\}$, where $J$ is the number of joints, and $\mathbf{p}_j \in \mathbb{R}^3$ is the axis-angle representation of the rotation of joint $j$.
The point cloud of the reference chair is denoted as $\mathcal{O} \in \mathbb{R}^{N\times3}$ and, likewise, the target chair as $\mathcal{O}^\prime \in \mathbb{R}^{N^{\prime}\times 3}$, where $N$ and $N^{\prime}$ are the number of points in the respective point clouds.
\par
The main question we seek to address in this section is the following: Given a person's sitting pose on a chair of \textit{known} geometry, how should this pose be adapted to a target chair with a previously unknown and different geometry? 
The target geometry could differ in several aspects, such as the height, type of armrest, or back support. 
Given a reference pose, $P$, of a character sitting or lying on the reference chair or sofa, $\mathcal{O}$, our goal is to optimise the target pose, $P{^\prime}$, for a previously unseen target object $\mathcal{O}^\prime$. 
We formulate this as a descriptor-matching problem on 3D points around the object, which allows us to reduce the dataset size significantly. 
Specifically, we train a neural network to estimate an $\mathsf{SE(3)}$-equivariant descriptor field around a given 3D object~\cite{simeonov2022neural}. 
$\mathsf{SE(3)}$-equivariance allows us to find a plausible orientation of the goal pose that does not penetrate the back of the chair.
Once trained, we can construct a descriptor, $\mathbf{Z}(\mathcal{O}, {P})$, for the \textit{skeleton} by aggregating the learned \textit{point} descriptors on the 3D keypoint locations of the reference pose.
These descriptors encode the spatial relationship between the character pose and object geometry. 
This is used to finally optimise for $P^{\prime}$ by minimising the discrepancy between $\mathbf{Z}(\mathcal{O}, P)$ and $\mathbf{Z}(\mathcal{O}^\prime, P^{\prime})$.
It is worth noting that such an approach does not require any manual labelling at inference but only a goal reference pose which can be conveniently sampled from the dataset or selected by the user, in contrast to prior work~\cite{starke2019neural,zhang2022couch}.
%
%
%
%
%
\begin{figure}[!t]
	\includegraphics[width=\linewidth]{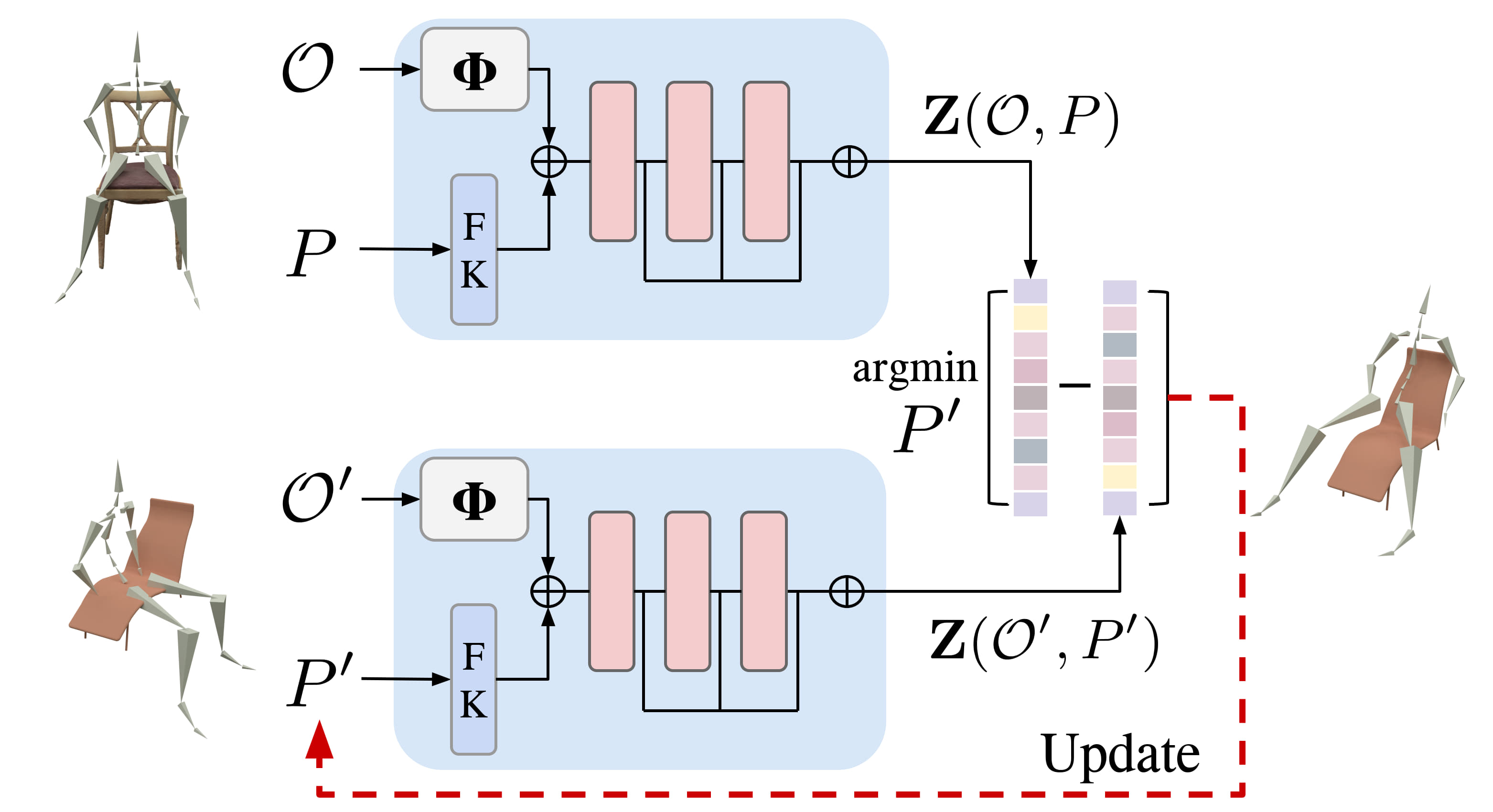}
	\caption
	{
            Goal Pose Synthesis (GPS) module. For the given reference object-pose pair and a target object, we generate neural descriptors $\mathbf{Z}(\mathcal{O}, P)$ and $\mathbf{Z}(\mathcal{O}^{\prime}, P^{\prime})$ that represent the relative geometry of the object and the pose, thereby allowing us to optimise the target pose $P^{\prime}$ (Sec~\ref{sec:gps}).
	}
	\label{fig:neural_shape_net}
 \vspace{-0.6cm}
\end{figure}
%
%
\subsubsection{Neural Descriptor Fields (NDF)}
Inspired by NDF~\cite{simeonov2022neural}, we learn a neural shape descriptor, $\Psi$ on objects belonging to the chair and sofa categories of the ShapeNet dataset~\cite{chang2015shapenet}. 
$\Psi$ is modelled as a continuous coordinate-based MLP, which maps a 3D point $\mathbf{x} \in \mathbb{R}^3$ to $\mathbf{z}_\mathbf{x}$, conditioned on an input object $\mathcal{O}$ as:
%
\begin{equation}
\mathbf{z_\mathbf{x}} = \Psi(\mathbf{x}, \mathcal{O}).
\end{equation}
%
As shown in Fig.~\ref{fig:neural_shape_net}, 
$\Psi$ 
consists of a PointNet~\cite{qi2017pointnet} encoder $\Phi$ and an occupancy network~\cite{mescheder2019occupancy} $\Omega$. 
%
$\Phi$ encodes the object $\mathcal{O}$ into a compact latent code that informs the occupancy network of the object geometry.
%
$\Omega$ is conditioned on the object's latent code and a query point, $\mathbf{x}$, which allow it to  
predict the occupancy flag at the query point:
%
\begin{equation}
\Omega(\mathbf{x}, \Phi(\mathcal{O})) \implies [0, 1].
\end{equation}
%
Since NDF's architecture is based on VectorNeurons~\cite{deng2021vector}, the descriptor is by design $\mathsf{SE(3)}$-equivariant.
As the $\Omega$ network is trained to estimate the occupancy values, the intermediate features implicitly model object-centric boundary information and spatial relationships between the points. 
Further, since $\Psi$ is a coordinate-based MLP, it allows us to extract these object-related features at an arbitrary point in space.
To obtain the descriptor for a point $\mathbf{x}$, the activations from $L$ layers of $\Omega$ are concatenated so that it contains hierarchical features:
%
\begin{equation}
 \mathbf{z}_{\mathbf{x}} = \oplus_{i=0}^L \Omega^i(\mathbf{x}, \Phi(\mathcal{O})). 
\end{equation}
%
By matching the point descriptors in two different objects, one can find correspondences between them. 
\par 
However, this descriptor, 
in the described form, 
only enables to rigidly match a coordinate frame at $\mathbf{x}$ from a source object to a target one, whereas we are interested in transferring the skeletal pose interacting with a source object to a target one.
Therefore, we introduce our skeletal pose descriptor (SPD) in the following.
%
%
\subsubsection{Skeletal Pose Descriptor (SPD)}
Our core idea is to aggregate a neural descriptor {$\Psi$} 
by querying all kinematic joints, which can be defined as
%
\begin{equation}
 \mathbf{Z}(\mathcal{O},P) = \{\mathbf{z}_\mathbf{x} \vert \mathbf{x} \in \mathcal{F}(P)\}, 
\end{equation}
%
where $\mathcal{F}(\cdot)$ is the differentiable forward kinematics function that outputs the set of 3D joint positions for a given pose.
However, constructing $\mathbf{Z}$ by querying $\Psi$ at only the 3D joint positions makes it sensitive to the descriptor's quality at that specific point.  
To make it robust to such outliers, we also sample the query points from the neighbourhood of each joint position.
Specifically, the points are sampled from a Gaussian distribution with standard deviation $\sigma$ centred at the joint positions $\mathcal{F}(P)_j$:
%
\begin{equation}\label{eq:pose_descriptor}
 \mathbf{Z}(\mathcal{O}, P) = \{\mathbf{z}_\mathbf{x_j} \vert \mathbf{x_j} \sim \mathcal{N}(\mathcal{F}(P)_j, \sigma)\}_{j=0}^J.
\end{equation}
%
Note that our descriptor $\mathbf{Z}$ is now a function of the skeletal pose $P$. 
Moreover, as each joint descriptor $\mathbf{z}_\mathbf{x}$ is $\mathsf{SE(3)}$--equivariant, it follows that our pose descriptor $\mathbf{Z}$ is $\mathsf{SE(3)}$--equivariant. 
In practice, this means that our method is robust to any rigid transform applied to the object, which is in contrast to previous work~\cite{starke2019neural,taheri2022goal} that assumes orientation and  translation of the object are known in advance. 
%
%
\subsubsection{Goal Pose Optimisation}
So far, we constructed a descriptor that is robust to rigid transforms and shape variations.
However, ultimately we are interested in matching a pose $P$ from a source object $\mathcal{O}$ to an unseen target object $\mathcal{O}^{\prime}$.
For this, we use the 3D skeleton pose descriptors $\mathbf{Z}$ defined above to find the optimal pose $P^{*}$ for $\mathcal{O}^{\prime}$.
Recall that our forward kinematics function $\mathcal{F}(P)$ in \eqref{eq:pose_descriptor} is differentiable, thereby allowing any gradients from the descriptor to also back-propagate into $P$.
\par
We, thus, formulate pose optimisation as an energy minimisation problem with the underlying postulate that the descriptors encode the geometric relationship between the pose $P$ and the object point cloud $\mathcal{O}$:
\[
E_d =  \sum_{j=1}^J \omega_j \big\lvert \mathbf{Z}(\mathcal{O}^{\prime}, P^{\prime})_j - \mathbf{Z}(\mathcal{O}, P)_j \big\rvert. 
\]
$\omega_j$ is a weight assigned to each joint descriptor.
For more details, please refer to Sec.~\ref{sec:implementation} in the supplementary material. 
\par%
Although the descriptor matching term, $E_d$, acts as the main data term driving the optimisation, it is limited in that it does not encode the human-body specific priors in it. 
After all, the descriptor is based on an occupancy network trained on a dataset of chairs and couches.
Therefore, we introduce additional regularisation terms to keep $P^{*}$ from drifting away from the manifold of plausible poses.
Concretely, we introduce an angle-limit constraint ($E_a$), a foot-floor penetration regulariser ($E_f$), and a pose regulariser ($E_p$).  
Our overall objective function is defined as: 
%
\begin{equation}
 P^* = \argmin_{P^{\prime}} 
 \bigg( E_d
 + 
\lambda_a {E}_a  
 +
 \lambda_f {E}_f  
 +
 \lambda_p {E}_p  
 \bigg). 
 \label{eq:objective_function} 
\end{equation} 
%
\par \noindent \textbf{Angle Constraints.} 
The SPD term, $E_d$ can result in unnatural twists between joints when source and target geometries vary significantly.
Similar to~\citet{shimada2021neural}, we penalise joint angles that are beyond a fixed bound:
\begin{equation}
 E_a = \lambda_a\sum_j^J \mathcal{A}({P'_j}), \,\text{with}
\end{equation}
\begin{equation}
\mathcal{A}({P'_j}) =
    \begin{cases}
        (P'_j - A_{j, max})^2, & \text{if } P'_j > A_{j, max} \\
        (A_{j, min} - P'_j)^2, & \text{if } P'_j < A_{j, min} \\
        0 & \text{otherwise.} 
    \end{cases}
\end{equation}
$P'_j$ is the joint rotation (Euler angles) and $A_{j, max}, A_{j, min}$ denote the upper and lower bounds of a joint rotation.
\par \noindent \textbf{Floor Penetration.} 
The floor penetration term penalises the penetration distance from the ground plane to all joints that are
below the ground and is formulated as: 
\begin{equation}
 E_f = \lambda_f \sum_j^J \lVert (\min(\mathcal{F}(P_j).y,  \mathcal{G}_y) - \mathcal{G}_y) \rVert^2. 
\end{equation}
$\mathcal{F}(P)_j.y$ denotes the y-magnitude of joint $j$ and $\mathcal{G}_y$ is the height of the ground. 
\par \noindent \textbf{Pose Regularisation.} 
Note that for symmetric objects, there can be more than one location with similar descriptor features. 
For example, consider a left-right symmetric chair.
The features on the right leg of the chair can be very similar to the features on the left leg due to similar geometric patterns and spatial relationship with the rest of the chair. 
Without further constraints, using an L1-loss in isolation would give us non-unique goal poses, which fall inside the local minima of the energy landscape. 
Additionally, we observe the optimisation may converge to an implausible sitting pose if the reference pose is very challenging (see Fig.~\ref{fig:ablation}-(c)). 
Thus, we introduce a pose regularisation term $E_r$ below that ensures that the target pose does not deviate drastically from the reference pose in the canonical space:
\begin{equation}
 E_r = \lambda_r  \lVert \mathcal{F}(P_{c}^{\prime}) - \mathcal{F}(P_{c}) \rVert^2,
\end{equation}
where the subscript $c$ denotes the canonicalised poses after aligning their root translation and orientation.
However, we set the weight $\lambda_r$ reasonably low for all experiments as a large weight can inhibit pose adaptation to novel shapes.
%
%
\subsection{Interactive Motion Generation} \label{sec:motion_net}
After having obtained $P^*$, 
we synthesise transitions from a start (standing) pose to the goal pose. 
To that end, we use a motion generation network which autoregressively synthesises idling, walking, sitting and transitioning motion between the three actions. 
Inspired by Starke \textit{et al.}~\cite{starke2019neural}, we 
propose a \textit{lightweight} Neural State Machine (\textit{l}-NSM). 
While NSM \cite{starke2019neural} defines interaction goal as a 6D vector encoding the goal position and orientation of the character's root, such an approach does not scale to our setting, which requires a full-body, 75DoF goal pose. 
Furthermore, we handle goal pose conditional object interaction (sitting down or lying down) and non-interaction locomotion without goal pose (walking or idling) in a single framework. 
Finally, since our goal pose is already object-aware, we do not use the environment and interaction sensors, thereby making the \textit{l}-NSM \textit{lighter}. 
As a result, our \textit{l}-NSM consists of the frame encoder and the goal encoder. 
\par 
%
The frame encoder encodes the character's current pose and trajectory within the $[-1s, 1s]$ time window $\mathcal{M}^t$ in an encoding $\Phi_f$ (in the following, we omit the superscript $t$). 
Specifically, it encodes the character's joint positions $\M_p \in \mathbb{R}^{3J}$, joint rotations $\M_q \in \mathbb{R}^{6J}$ and the linear 
 joint velocities $\M_v \in \mathbb{R}^{3J}$. 
It also takes as input the root trajectory information such as the ground-plane projections of the root, $\M_r \in \mathbb{R}^{2L}$, for the past and future $L$ frames as well as the corresponding 2D orientation vectors $\M_o \in \mathbb{R}^{2L}$. 
Finally, in encodes the soft action label $\M_a \in \mathbb{R}^{4}$ corresponding to idle, walking, sitting and standing actions.
%
The goal encoder $\Phi_g$ encodes 
the target goal pose $\mathcal{G}^t$. 
The input to $\Phi_g$ includes the goal-pose's joint positions $\mathbf{G}_p$ as well as the ground-truth goal trajectory $\mathbf{G}_r \in \mathbb{R}^{3L}$ and orientations $\mathbf{G}_o \in \mathbb{R}^{3L}$.
$L$ is sampled from the future $[0, 2s]$ time window.
Similar to $\Phi_f$, $\Phi_g$ also encodes the action labels $\mathbf{G}_a$ but with one-hot encoding. 
At test time, $\mathbf{G}_r$ is the same as $\mathcal{F}(P^{*})$.
%
Let $\mathbb{M}_c = (\M_p, \M_q, \M_v)$, 
$\mathbb{M}_{ro} = (\M_r, \M_o)$, 
$\mathbb{G}_{roa} = (\G_r, \G_o, \G_a)$.
We then have
\begin{align}
    \Phi_f : \mathbb{R}^{|\mathcal{M}^t|} \rightarrow \mathbb{R}^{d}, & \quad \mathcal{M}^t = (\mathbb{M}_c, \mathbb{M}_{ro}, \M_a),  \\
    \Phi_g : \mathbb{R}^{|\mathcal{G}^t|} \rightarrow \mathbb{R}^{d}, & \quad \mathcal{G}^t = (\delta\G_p, \mathbb{G}_{roa}), 
\end{align}
where $\delta$ indicates whether the character is in locomotion mode ($\delta=0$) or interaction mode ($\delta=1$). 
Given the two encoders projecting the character's current state and the target state into a $d$-dimensional latent space, the \textit{l}-NSM decoder, $\Psi : \mathbb{R}^{2d} \rightarrow \mathbb{R}^{|{Y}^{t+1}|} $, then decodes them into the motion for the next frame with: 
\begin{equation}
\small
Y^{t+1} = \bigg(\mathbb{M}_c, \hat{\mathbb{M}}^{t+1}_c, \mathbb{M}^{t+1}_{ro}, \hat{\mathbb{M}}^{t+1}_{ro}, \mathbb{G}^{t+1}_{roa}, c^{t+1}, f^{t+1}\bigg). 
\end{equation}
%
%
$\hat{\mathbb{M}}_c$ denotes the inverse pose defined relative to the goal coordinate system. 
Likewise, $\hat{\mathbb{M}}_{ro}$ denotes the inverse, goal-centric trajectory.
$c_t \in \{0, 1\}$ are the floor-contact predictions which, if $1$, trigger the post-processing inverse-kinematics routine to reduce footsliding.  
The decoder also estimates the phase $f_t \in \mathbb{R}^L$ of the gait which is fed into the gating network outputting the blending coefficients of the experts. 
We refer the reader to~\citet{starke2019neural} for details. 
\par \textbf{Bidirectional Pose Blending.} 
Handling unconditional locomotion and goal-conditioned transitions in an integrated architecture is challenging. 
Simply conditioning \textit{l}-NSM on $P^{*}$ proves to be insufficient to accurately reach it. 
In practice, we observe that the network generates a series of poses resembling the training sequence instead of 
reaching the unseen goal poses provided by GPS. 
Moreover, switching the animation into interaction mode 
often leads to sudden instabilities in the generated motion, as the transition from locomotion mode to interaction mode is abrupt. 
To address these artefacts, we adapt the bidirectional control routine proposed by~\citet{starke2019neural} to our setting. 
The core idea of bidirectional control is to blend the ego-centric forward motion $\mathbb{M}_c$ and trajectory $\mathbb{M}_{ro}$ with the estimated goal-centric inverse motion $\hat{\mathbb{M}}_c$ and $\hat{\mathbb{M}}_{ro}$.
These two predictions should ideally lead to the same update after transformation to the world space but often differ in inference due to autoregressive error accumulation and time needed to adjust to user input. 
Therefore, the forward and the reverse poses (and trajectories) are blended using: $\mathbb{M}^{\prime}_p = \mathbb{M}_p + \lambda_b T \hat{\mathbb{M}}_p$, where $T$ is the transformation matrix between the goal and ego-centric coordinate systems, and $\lambda_b$ is the blending weight derived from the action prediction and distance to goal (see the supplementary material). 
\par
Recall that our goal pose, $P^{*}$ is automatically computed by the GPS module.
This requires us to define a per-joint goal coordinate system where the origin lies at the joint's position, which we call the goal-joint coordinate system during object interaction.
The orientation is computed by recovering the normal of the torso-plane of $\mathbf{P}^{*}$. 
As a result, the proposed Bidirectional Pose Blending elegantly generates seamless object interaction and locomotion.
It is worth noting that we do so with minimal overheads to already large input and output dimensions of \textit{l}-NSM. 
Once the goal pose is computed, \textit{l}-NSM synthesises motion in real time and allows both low-level interactive control and high-level goal-conditioned motion synthesis. 
%

%
%
\begin{figure}[!t]
	\includegraphics[width=\linewidth]{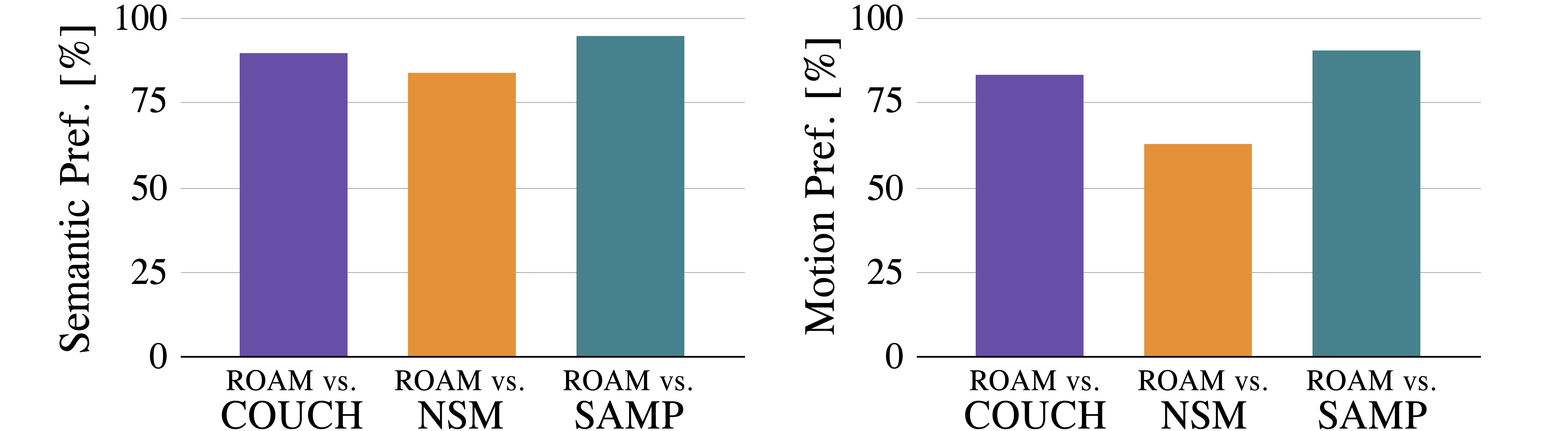}
	\caption
	{
    Results of the perceptual study for qualitative comparison of ROAM against the state of the art. 
    Our method is consistently preferred over all other approaches, in terms of, both, semantic coherence with the chair's geometry, and motion realism.
	}
	\label{fig:perceptual_study}
 \vspace{-0.6cm}
\end{figure}
\section{Experiments}\label{sec:experiments}
We compare our ROAM to its most related approaches, i.e, NSM~\cite{starke2019neural}, SAMP~\cite{hassan2021stochastic} and COUCH~\cite{zhang2022couch} and evaluate it in terms of the animation realism and the plausibility of the synthesised sitting poses. 
Due to different skeleton definitions, types of interaction actions and number of reference objects, we retrain the models on our dataset. Please refer to the supplementary material Sec.~\ref{sec:dataset} for more details.
NSM needs to be provided with a point of contact labels and the object orientations; COUCH assumes a predefined chair orientation), while we do not make any of such assumptions. 
All the methods are provided a character's starting pose and orientation along with the point cloud of the chair in the scene. 
All of them receive the same initial state of the character and the environment. 
The evaluation settings and their corresponding metrics are discussed next.
For detailed visual comparisons with the state-of-the-art methods, the reader is referred to the supplementary video. 
%
%
\begin{figure*}[!t]
	\includegraphics[width=\textwidth]{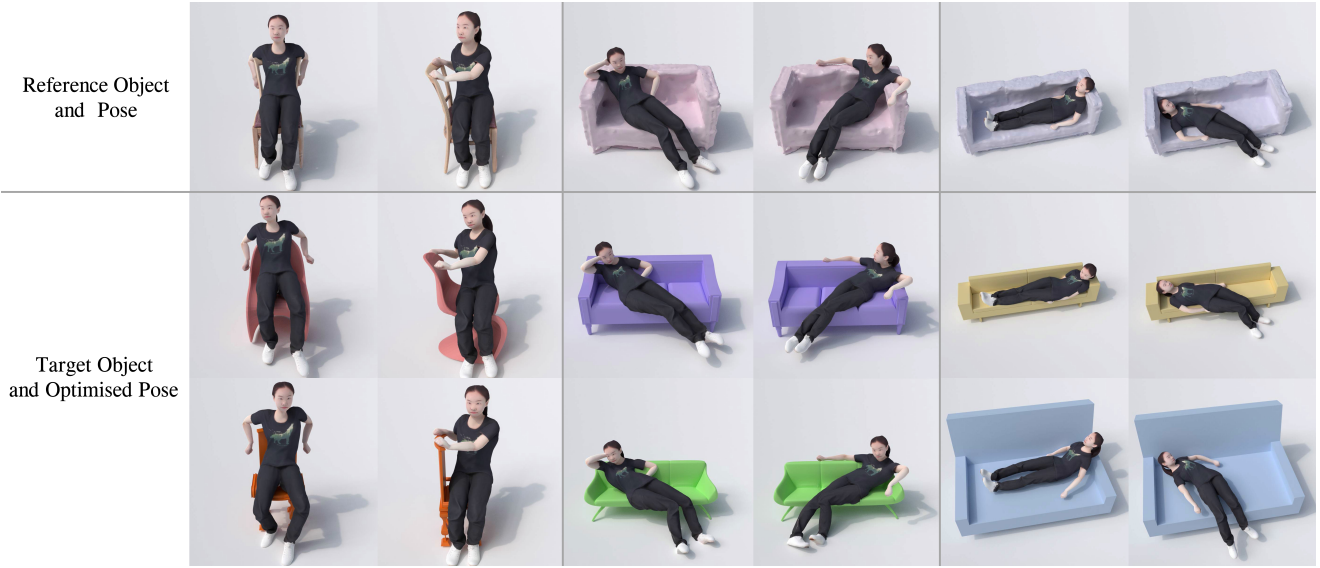}
	\caption
	{
	    Qualitative results of the Skeletal Pose Descriptor-based goal pose optimisation (Sec.~\ref{sec:gps}). 
            Given a reference pose for a reference chair (top row), our method can adjust the reference pose to an unseen chair (second and third row).
            We show a variety of poses in the range from sitting on chairs (first column), sitting on sofas (middle column) and lying on sofas (last column).   
            Please watch our supplementary video for animations of the optimisation process.
	}
	\label{fig:qualitative}
 \vspace{-0.4cm}
\end{figure*}
%
%
\par
\subsection{Perceptual User Study}\label{ssec:perceptual_study}
Recall that NSM, COUCH and SAMP animations are conditioned on the points of contact alone and often do not adapt well to object geometries.
Such cases are hard to numerically account for, and hence, necessitate a perceptual user study to evaluate the quality of motion synthesis. 
The participants are asked to compare $12$ randomly generated animation sequences in a forced-choice manner, \textit{i.e}, the participant must choose the best motion sequence out of the presented sequences. 
We ask the participants the following two questions for each set of motions: \textit{Which motion do you prefer in terms of the naturalness of the motion? (Realism)} and \textit{Which motion do you prefer in terms of the agreement of the sitting motion with the chair geometry? (Semantics)}
\par
\noindent \textbf{Results:}
We received responses from $50$ participants, leading to $600$ valid comparisons in total ($200$ per competing method); Fig~\ref{fig:perceptual_study} summarises the results. 
Our method is preferred over NSM, COUCH and SAMP by a significant margin.
We calculate $p$-values (binomial test) for our comparison with each method and observe statistically significant results with $p{<}0.001$ for all three methods.
Interestingly, NSM's motion generation performs better than the other two more recent methods in terms of user preference.
It is worth noting that NSM chairs have been hand-labelled for the goal position whereas our motion synthesis relies on the automatically generated goal pose.
%
\subsection{Quantitative Results} \label{sec:quantitative}
We next perform quantitative analysis to evaluate our design choices. 
However, establishing a firm evaluation metric for our setting is challenging as no single one would account for all aspects. 
\par
\noindent \textbf{Goal Distance Error.} We evaluate whether the motion synthesis module reaches the intended target by measuring the Position Error (PE) to the goal.
The comparison is done by using 100 unseen test objects with the character starting position randomly sampled between two to four meters to the goal object with a starting angle between -30\degree and 30\degree. We record the minimum goal error reached within 300 frames in \cref{tab:goal_error}.
For comparison with NSM, we consider the manually labelled target position on the chair or couch as the goal position and compute the distance of the hip joint to it (Hip PE in \cref{tab:goal_error}). 
For SAMP, we use their generated hip positions and directions.
Next, since COUCH facilitates reaching a goal pose based on the hand contacts, we use the generated hand contact positions and measure their distance from the hand positions in the synthesised sitting motion (Hand PE in \cref{tab:goal_error}).
In the above setting, we choose the `hip' joint and the `hand' joints of ${P}^{*}$ as our target positions, respectively. 
We also present an ablation of our method without bidirectional blending, which proves to be inferior.
Finally, we evaluate our full-body per-joint position error between the goal pose ${P}^{*}$ and the \textit{l}-NSM output (Avg PE in \cref{tab:goal_error}).
\begin{table}
\begin{center}
\small
\begin{tabular}{lrrr}
    \toprule
    Method  & Hip PE$\downarrow$  & Hand PE$\downarrow$   & Avg PE$\downarrow$ \\
    \midrule
    
    NSM	\cite{starke2019neural}  &     2.01          &      -     &          -   \\ 
    SAMP \cite{hassan2021stochastic}  &     1.86            &      -     &         -     \\ 
    COUCH \cite{zhang2022couch} &             10.12   &   10.47       &         -      \\ 
    \textbf{Ours (No Blending)}	 &   7.30    &  21.37       & 14.31          \\
    \textbf{Ours}	                &   \textbf{1.33}    &  \textbf{4.12}       & \textbf{2.73}          \\
    \bottomrule
    \end{tabular}
    \end{center}
 \caption
    {
    Position Error (PE) Comparison with the state-of-the-art methods on the ability to accurately reach the goal. 
    For us, we consider the goal pose's hip position and hand positions as the target to compute Hip PE and Hand PE.
    For COUCH, the generated hand contact labels are considered the target with our annotated goal orientation. 
    For SAMP, we use the predicted hip position and direction as the goal.
    Finally, for NSM we compute the error with respect to the manually labelled target goal position.
    }   
 \label{tab:goal_error}	
 \vspace{-0.6cm}
\end{table}
%
%
%
%

\begin{figure*}[!ht]
	\includegraphics[width=\linewidth]{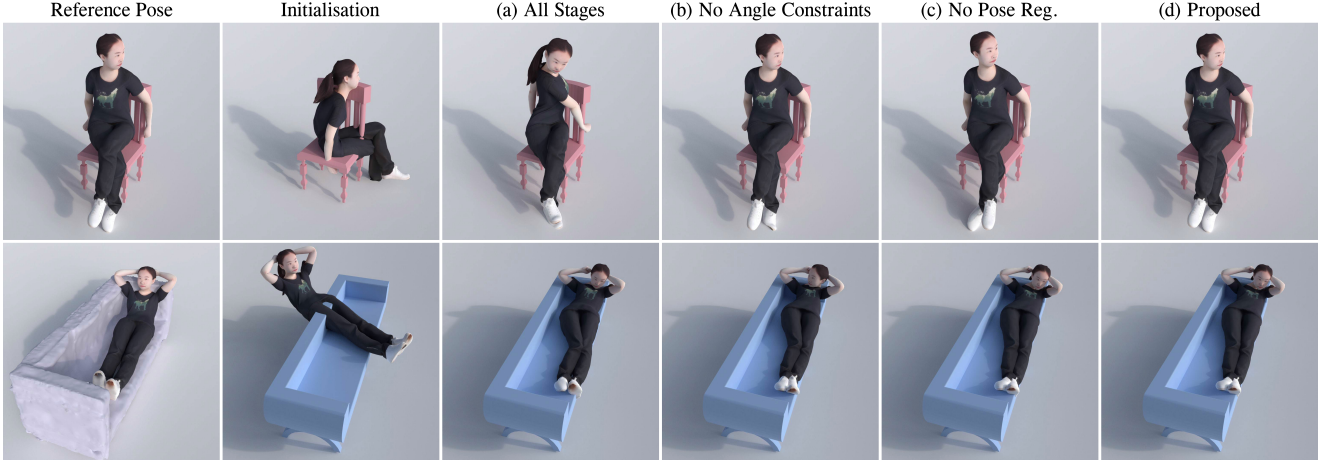}
	\caption
	{
	    We perform three ablation experiments for goal pose optimisation. (a) \emph{All Stages} optimises all DoF simultaneously with all energy terms added from the first iteration. This strategy leads to twists in body joints. (d) Our proposed strategy with three-stage optimisation, ${E}_a$ and ${E}_r$. We empirically find this combination leads to the best qualitative results. 
	}
	\label{fig:ablation}
    \vspace{-0.3cm}
\end{figure*}
%
%
\subsection{Qualitative Results} \label{sec:qualitative}
In Fig.~\ref{fig:qualitative}, we show how GPS adapts the reference pose to the target geometry for sitting on a chair and sofa and lying on a sofa.
\par
\begin{figure}[!hbt]
    \centering
    \includegraphics[width=\linewidth]{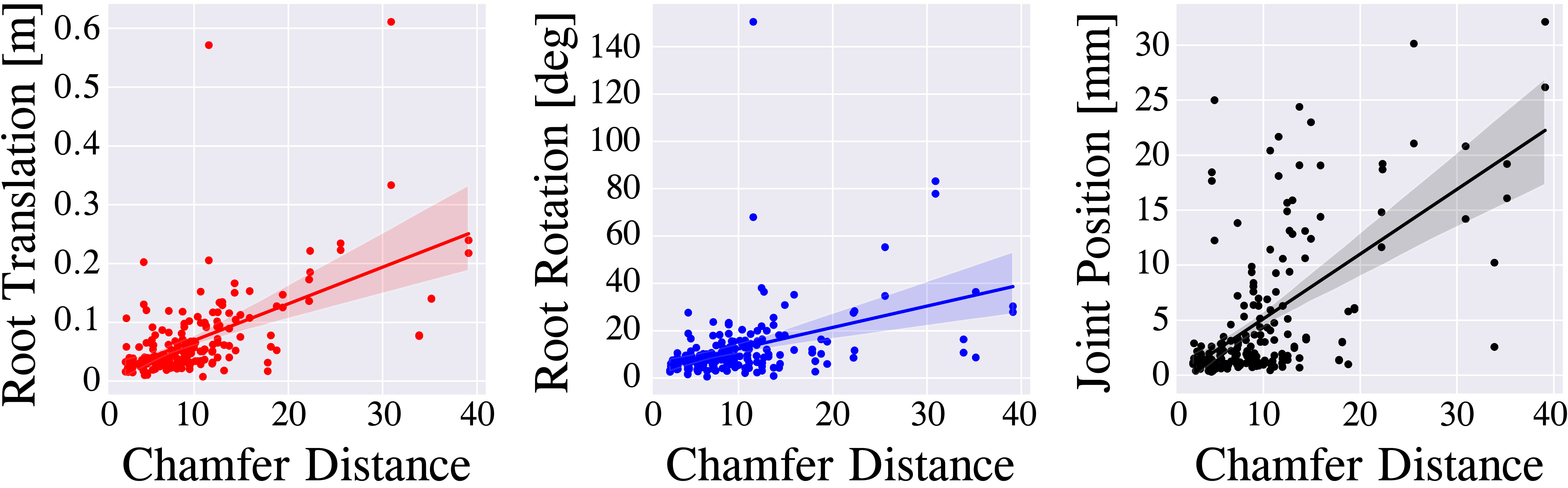}
    \caption{A plot depicting the difference between the optimised goal pose and the reference pose as the Chamfer distance between the target couch and the reference couch changes across the test couches. For each optimised goal pose $P^{*}$, we measure the change in the root translation (left), root orientation (centre), and the root-normalised joint position error (right) with respect to the initial reference pose. We observe that $P^{*}$ remains robust to large variations in the target object's geometry, as measured by the Chamfer distance. 
    }
    \label{fig:scatter}
    \vspace{-0.6cm}
\end{figure}

Another interesting analysis is to measure the difference between the optimised novel pose and the reference pose as the target chair becomes more distant from the reference chair in terms of the Chamfer distance.
To perform this analysis, we align the position and orientation of $100$ chairs in the ShapeNet dataset \cite{chang2015shapenet} and use a single reference pose to generate goal poses for each chair.
We then compute the difference between the reference and the target pose in terms of their root translation, root orientation and root-aligned 3D joint positions.
The scatter plot in \cref{fig:scatter} follows the expected trend and we observe that the goal pose has to adjust significantly as the target chair's geometry deviates from that of the reference chair.
\subsection{Ablation} \label{sec:ablation} 
In Fig.~\ref{fig:ablation}, we present ablation visualisations of three design choices for two poses of different difficulty.
In the first ablation, we discard the proposed sequential three-stage optimisation routine (see Sec.~\ref{sec:implementation} in the supplementary material) and instead jointly optimise for the root orientation, root translation and root-relative articulations. 
We notice that doing so often leads to sub-optimal results because in the beginning of the optimisation, the reference pose can be in a significantly different orientation and the corresponding gradients are expected to be uninformative for optimising joint articulations.
Next, we evaluate the effect of adding angle constraints ${E}_a$ in our loss function in \cref{eq:objective_function}.
${E}_a$ penalises implausible joint configuration and leads to more natural results, especially for challenging poses and out-of-distribution chairs.
We also evaluate the choice of including the pose regulariser ${E}_r$ on the joint positions in the canonical space. 
This regulariser becomes important to ensure that the optimisation does not stuck in local minima and often leads to more natural poses. 
Finally, the rightmost column of Fig.~\ref{fig:ablation} shows that the proposed design choices lead to improved results. 
%

%
%
\section{Discussion and Conclusion} \label{sec:conclusion}
We presented ROAM, a framework for robust synthesis of human-object interactions. 
Focusing on chairs and couches as an important instance arising in many applications, we demonstrated that the proposed two-step strategy of ROAM allows robust generalisation to a variety of unseen objects.
Furthermore, we showed that the proposed bidirectional pose blending not only allows the motion synthesis pipeline to smoothly transition to the goal pose but also generates natural motion that the users found to be better than existing scene interaction methods.
Crucially, we could achieve this by training with motion capture data involving only a \emph{single} chair/couch exemplar. 
However, our method leads to artefacts when the target geometry differs significantly from the source geometry.
We believe these artefacts can be removed by leveraging character mesh instead of the skeleton representation with physical plausibility constraints (see supplementary material for more discussion on future work). 
With the ever-growing demand for realistic 3D characters in virtual worlds, we envision our principled and practical approach to contribute to the creation of scalable immersive visual experiences.

\vspace{0.2cm}

\noindent{\bf Acknowledgement}: This work was carried out as part of a dissertation at Saarland University. This project was also supported by Saarbrücken Research Center for Visual Computing, Interaction and AI. Christian Theobalt was supported by ERC Consolidator Grant 4DReply (770784). We thank Janis Sprenger for helpful discussions on experiment design and visualisation.

{
    \small
    \bibliographystyle{ieeenat_fullname}
    \bibliography{main}
}

\clearpage
\maketitlesupplementary
\appendix
The appendix contains information on the dataset (Sec.~\ref{sec:dataset}), $\mathsf{SE(3)}$-equivariant property of the proposed method (Sec.~\ref{sec:se3}), implementation details (Sec.~\ref{sec:implementation}) and and quantitative ablations (Sec.~\ref{sec:quantitative_ablation}) on the point query strategy. 
In addition, we also include a more detailed discussion on the limitations and future work. (Sec.~\ref{sec:limitations}).
\section{Dataset} \label{sec:dataset}
To train our approach, we recorded a new dataset of 
two subjects performing idling, standing and walking, and one subject interacting with objects: eg. sitting on a chair and a sofa and lying down on a sofa in a multi-view green screen studio.
Notably, we only record motions for a single object per action category (one chair and one sofa for sitting, and one sofa for lying down).
In total, we recorded around 90 minutes of motion at a frame rate of 25fps using 30 calibrated and synchronised RGB cameras.
To recover the skeletal motion, we leverage a markerless motion capture software~\cite{captury}.
Foot contact and phase labels are automatically annotated based on the altitude of the foot joints. 
We will release our dataset for future research.
\section{Robustness to Orientation Initialisations} \label{sec:se3}
\begin{figure*}[!t]
	\includegraphics[width=\linewidth]{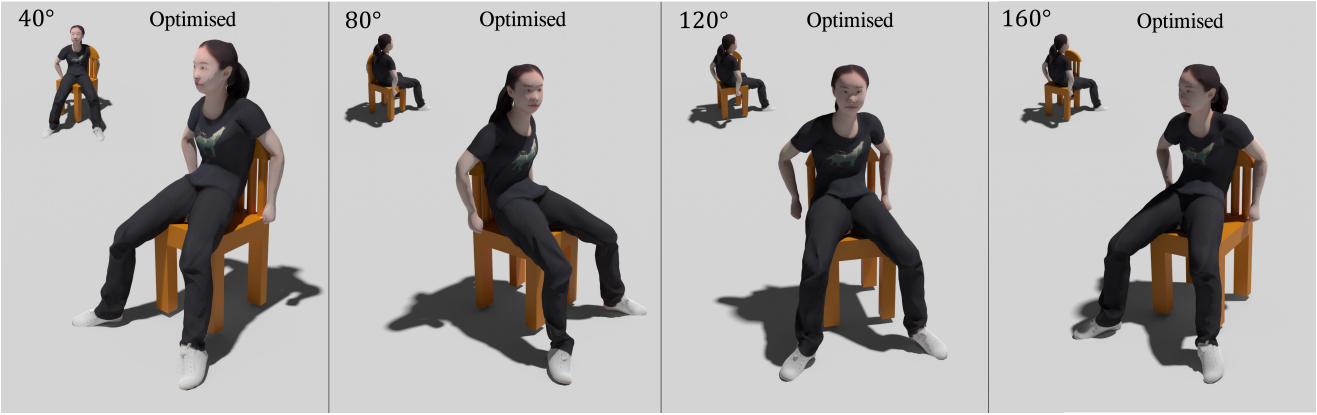}
	\caption
	{
    We show the robustness of the proposed Goal Pose Synthesis module to different initial root orientation initialisations due to the $\mathsf{SE(3)}$-equivariant property.
    For each column, the image at the top left shows the initialised reference pose 
    rotated 40\degree, 80\degree, 120\degree and 160\degree respectively on the target chair before the optimisation begins.
    GPS optimisation corrects the rotation error.
    }
	\label{fig:qualitative_rotation}
\end{figure*}

We show the robustness of our goal pose synthesis (GPS) module to perturbations in initial orientations in Fig~\ref{fig:qualitative_rotation}. 
As shown on the top left, our initial pose for optimisation is the same as the reference pose, with the root position shifted to the nearest chair point. 
Our method is robust even with poor initialisation (\textit{e.g.,} the initial pose is rotated 40\degree, 80\degree, 120\degree and 160\degree respectively). 
The SPD optimisation routine not only correctly optimises the pose orientation, but also adjusts the pose according to the height of the target chair.
%
%
%
%
%
%
\section{Implementation Details} 
\label{sec:implementation}
\par \noindent \textbf{SPD Optimisation}. 
We sample a query point cloud with $(\mu=0, \sigma=0.025)$ and centre it around all joint locations including eight additional virtual joints at the midpoint of the limbs.
We re-use the same query point cloud for obtaining the descriptor for reference pose ${P}$ and target pose ${P'}$.
The descriptor for each joint is concatenated in a fixed order for SPD optimisation.
The energy function we optimise is high-dimensional, non-linear, and contains local optima. 
Thus, na\"ively optimising over all the variables jointly typically leads to sub-optimal results.
To alleviate this problem, we perform optimisation in three sequential stages:
First, we only optimise the root translation and rotation, \textit{i.e.}, the rigid transform of the character.
In this stage, we empirically set $\omega_j$ for the knee descriptors as  $3.0$, while keeping it $1$ for the rest of the joints. 
Since knee positions are typically close to the chair's surface, the descriptors are more accurate compared to the joints which are far away.
They are also more indicative of the relative orientation between the sitting pose and the chair compared with the hip descriptor, which is also close to the chair surface.
Further, we do not activate $E_a$, $E_p$ and $E_f$ in this stage, thus $\lambda_a, \lambda_a, \lambda_f$ are set to 0. 
This stage takes 250 iterations.
2) Next, we jointly optimise all the DoF of the skeleton without using the energy terms.
We also reset all joints to equal weight. 
This stage takes 100 iterations.
3) In the final stage, regularisation is activated with the following weights: $\lambda_p=1\mathrm{e}{-2}$, $\lambda_a=1\mathrm{e}{-3}$ and $\lambda_f=1\mathrm{e}{-8}$.
The upper and lower bounds of joint angles in $E_a$ are statistically computed from our dataset and we also relax it by $\pm10\degree$ to allow novel poses to be generated on out-of-distribution target objects.
The final stage stakes 150 iterations and in total we optimise for 500 iterations.
We use Adam~\cite{Kingma2014AdamAM} optimiser with a learning rate of $1$ for root translation and $0.01$ for the rest of DoF. 
\par \noindent \textbf{\textit{l}-NSM} 
We train \textit{l}-NSM for 150 epochs with 10 experts in the decoder.
AdamW~\cite{Loshchilov2017DecoupledWD} optimiser is used with an initial learning rate of $1\mathrm{e}{-4}$ and with cosine annealing. 
At inference time, ROAM allows both low-level and high-level controls for the character to interact with the object. 
Low-level control enables users to drive the character using the keyboard after loading the goal poses. 
In high-level mode, the user only needs to specify the starting location and orientation and the interaction sequence will be automatically generated. 
To achieve this, we first set a walking goal at 0.3 meters away from the goal foot positions along the normal of the torso plane for sitting.
For lying down, the torso plane is not indicative of which direction the object should be approached.
Based on the observation that a person is likely to approach a sofa from the front side, a ray is cast from the goal hip joint along the difference vector between two goal foot positions.
A hit indicates that the ray is pointing to the back of the sofa; thus, the character should approach from the other side.
Bidirectional pose blending is only activated when the user gives an interaction signal in low-level mode, or when walking is completed in high-level mode.
We set a blending weight $\lambda_b$ which is a product of current interaction action value $\M_a^i$ (a scalar ranging between $0$ and $1$) and a distance-based weight defined as: 
$\lambda_b = \M_a^i \cdot (-\frac{1}{1 + e^{20 \cdot (-(d - 0.25))}} + 1)$ 
where $d$ is the average joint distance between the current and the goal poses.

\section{Quantitative Ablation} \label{sec:quantitative_ablation}
\begin{figure}
    \centering
    \includegraphics[width=0.95\linewidth]{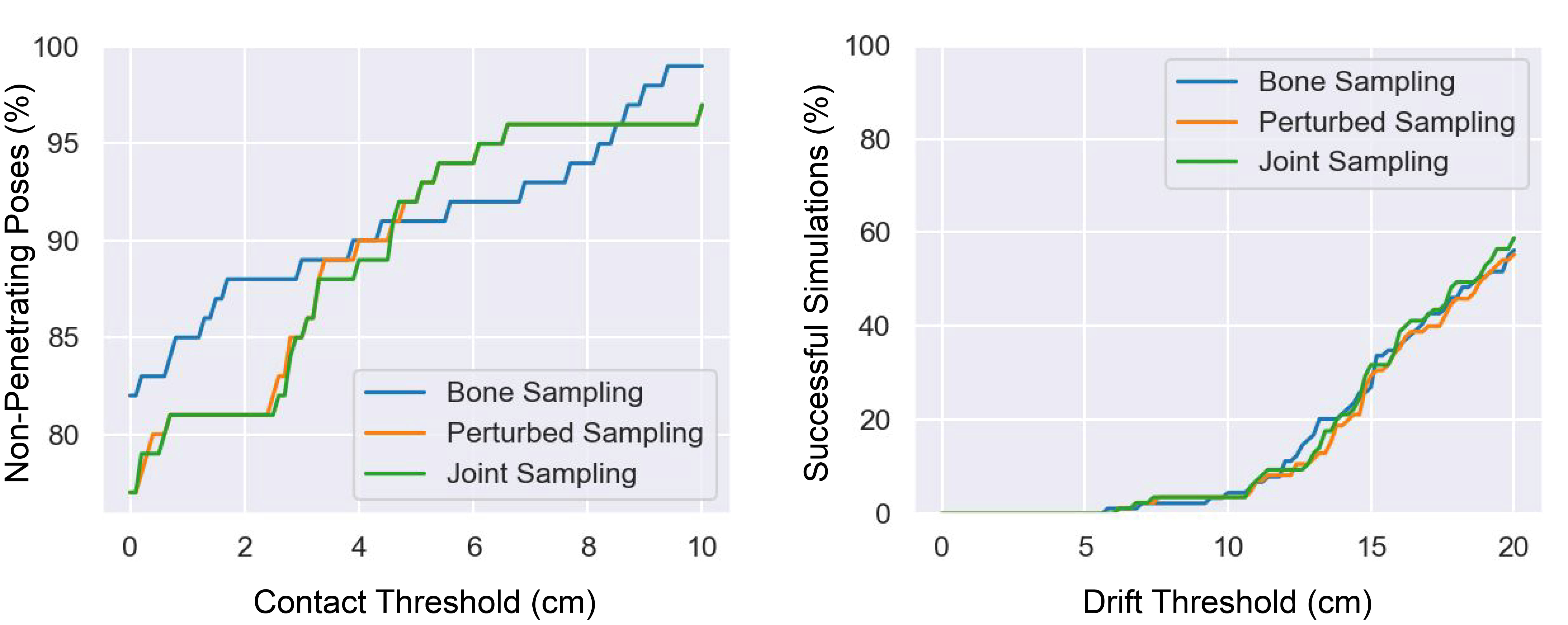}
    \caption{Comparison of the penetration AUC curves (left) and simulation success curves (right) of the ablations. }
    \label{fig:auc_penetration}
\end{figure}

We now discuss the analysis of our method and compare it with its ablated versions. 
For quantitative evaluations, we perform the following ablations. 
\textit{Joint Sampling} is our baseline where each query point cloud is centred at the exact joint positions. 
\textit{Bone Sampling} refers to adding eight additional query point clouds from the midpoint of the limbs. 
Note that we do not add more midpoints along the spine since they are relatively denser than limb joints.
\textit{Perturbation Sampling} refers to adding noise to each joint query point cloud to lift the optimisation out of potential local minima.
We first compute the pose plausibility by measuring the percentage of non-penetrating goal poses.
We do it in the PyBullet~\cite{coumans2016pybullet} physics simulator and plot the AUC curve of the percentage of non-penetrating poses against collision thresholds (\cref{fig:auc_penetration}(left)).
The contact collision threshold defines the degree of allowed penetration in order to account for the inaccuracies of the body and object modelling in the simulator. 
We also evaluate if the non-penetrating sitting poses lead to a stable sitting posture by running physics simulations on a humanoid robot with the same DoF as in our skeleton.
We let the humanoid simulate for $t=1s$ at a simulation rate of $240$Hz and measure the drift of the character before and after the simulation.
Again, we plot an AUC curve of the per-joint drift for several thresholds on the maximum allowed drift (\cref{fig:auc_penetration}(right)).
\par
We observe that sampling points along the bone leads to lesser penetration, but the success rate remains unaffected by the type of ablation. 
This is expected as sampling along the bone provides a larger coverage of the occupancy field around the chair, unlike the case where we sample around the 3D joints.

%
%
\section{Limitations and Future Work} \label{sec:limitations}
%
%
\begin{figure}
	\includegraphics[width=\linewidth]{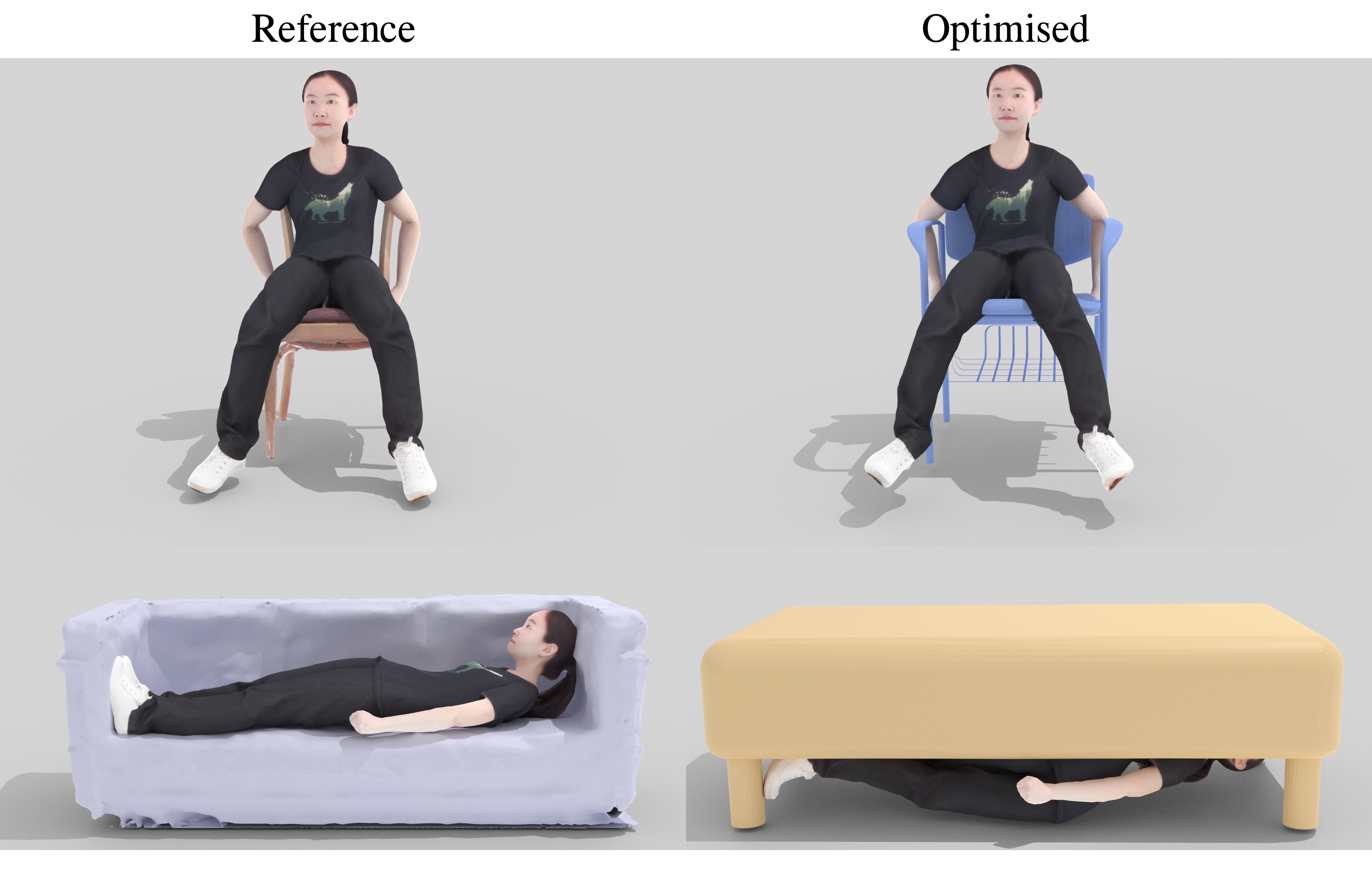}
	\caption
	{
	    Limitations. The top row shows the challenging scenario of adapting a pose to a target chair with armrests. Since the correspondences are not well-defined, our optimised pose has penetration at the elbow region. The bottom row shows that SPD optimisation leads to a physically implausible pose.
	}
	\label{fig:limitations}
	\vspace{-0.4cm}
\end{figure}
%
%
%
While we have highlighted experimentally the generalisability and data efficiency of our approach, it is not free from limitations. 

First, ROAM reasons on the level of a kinematic skeleton. 
Consequently, neither the human body shape nor the clothing is accounted for and we observe occasional character-object penetrations. 
Thus, a promising avenue for future work is to explicitly model the dense geometry of the virtual character and its deformable surface, from which both the goal pose estimation and the motion synthesis would benefit. 
Moreover, additional physics-based constraints \cite{PhysCapTOG2020} could further boost the realism of our results (\textit{e.g.,} by mitigating foot sliding, enforcing contacts and penalising implausible poses as shown in (Fig.~\ref{fig:limitations})). 
In addition, NDF for estimating the goal pose inherently exhibit only low-frequency variations. 
Therefore, only rather coarse-grained pose estimations can be performed (\textit{e.g.,} preventing hand articulations, which are, therefore, not part of our model). 
Furthermore, the quality and variability of our pose estimator are tightly linked to the NDF training dataset. 
Substantially out-of-distribution objects can, therefore, lead to 
unpredictable results. 
Note that we inherit these limitations from the original NDF formulation; any respective advances will directly benefit our method. 
In terms of motion generation, \textit{l-}NSM is scene-agnostic and cannot avoid obstacles along the way.
When an out-of-distribution goal pose is provided, the bidirectional pose blending scheme could lead to artefacts.
Finally, we have illustrated the feasibility of our approach using only two actions (sitting and lying) on two categories (chairs and sofas). 
Yet, no component of our method is specifically designed for this particular combination and it could be trained for other combinations once other datasets are available in future. 
%


\end{document}